\documentclass[letterpaper, 10 pt, conference]{ieeeconf}  

\IEEEoverridecommandlockouts                              
\usepackage{graphicx} 
\usepackage{epsfig} 
\usepackage{amsmath} 
\usepackage{amssymb} 
\usepackage{pifont}
\usepackage{soul,color} 
\usepackage{mathtools}
\usepackage{booktabs}
\usepackage{multirow}
\usepackage{balance}
\usepackage{gensymb}
\usepackage{tabulary}
\usepackage[numbers,sort&compress]{natbib}
\usepackage{dblfloatfix}
\usepackage{layouts}
\usepackage{algorithm}
\usepackage{algpseudocode}
\usepackage{lipsum}
\usepackage{subcaption}
\usepackage{siunitx}
\usepackage{hyperref}

\graphicspath{{figures/}}
\newcommand{\etal}{\mbox{{et al.\ }}}
\newcommand{\figGap}[0]{\vspace{-1.4\baselineskip}}

\newcommand{\diag}{\mathop{\mathrm{diag}}}

\title{\bf 
Learning Tactile Models for Factor Graph-based Estimation
\vspace{-2mm}
}
\author{Paloma Sodhi$^{1, 2}$, Michael Kaess$^{2}$, Mustafa Mukadam$^{1}$, and Stuart Anderson$^{1}$\\[2mm]
$^{1}$Facebook AI Research, $^{2}$Carnegie Mellon University
\thanks{\footnotesize{Work done when Paloma Sodhi interned at Facebook AI Research. {\tt psodhi@cs.cmu.edu}. All supplementary material can be found on {\tt https://psodhi.github.io/icra21tactile}}}
\vspace{-2mm}
}

\begin{document}

\maketitle
\thispagestyle{empty}
\pagestyle{empty}

\begin{abstract}
	We're interested in the problem of estimating object states from touch during manipulation under occlusions. In this work, we address the problem of estimating object poses from touch during planar pushing. Vision-based tactile sensors provide rich, local image measurements at the point of contact. A single such measurement, however, contains limited information and multiple measurements are needed to infer latent object state. We solve this inference problem using a factor graph. In order to incorporate tactile measurements in the graph, we need local observation models that can map high-dimensional tactile images onto a low-dimensional state space. Prior work has used low-dimensional force measurements or engineered functions to interpret tactile measurements. These methods, however, can be brittle and difficult to scale across objects and sensors. Our key insight is to directly learn tactile observation models that predict the \textit{relative pose} of the sensor given a pair of tactile images. These relative poses can then be incorporated as factors within a factor graph. We propose a two-stage approach: first we learn local tactile observation models supervised with ground truth data, and then integrate these models along with physics and geometric factors within a factor graph optimizer. We demonstrate reliable object tracking using \textit{only tactile} feedback for $\sim$150 real-world planar pushing sequences with varying trajectories across three object shapes.
\end{abstract}

\section{Introduction}
\label{sec:introduction}

We look at the problem of estimating object states such as poses from touch. Consider the example of a robot manipulating an object in-hand: as the object is being manipulated, it is occluded by the fingers. This occlusion renders visual estimation alone insufficient. \textit{Touch} in such cases can provide local, yet precise information about the object state.

The advent of new touch sensors \cite{lambeta2020digit, yuan2017gelsight, donlon2018gelslim, yamaguchi2016fingervision} has enabled rich, local tactile image-based measurements at the point of contact. However, a single tactile image only reveals limited information that may correspond to multiple object poses, making it hard to directly predict poses from an image. We hence need to be able to reason over multiple such measurements to collapse uncertainty. This results in the inference problem: \emph{Compute the sequence of latent object poses, given a stream of tactile image measurements.}

We solve this inference problem using a factor graph \cite{dellaert2017factor, dellaert2020factor} that offers a flexible way to process a stream of measurements while incorporating other priors including physics and geometry. The factor graph relies on having local observation models that can map measurements into the state space. Observation models for high-dimensional tactile measurements are, however, challenging to design. Prior work \cite{yu2018realtime, lambert2019joint} used low-dimensional force measurements or engineered functions for high-dimensional tactile force measurements. These, however, can be brittle and difficult to scale across objects and sensors.

\begin{figure}[!t]
	\centering
	\includegraphics[width=0.95\columnwidth]{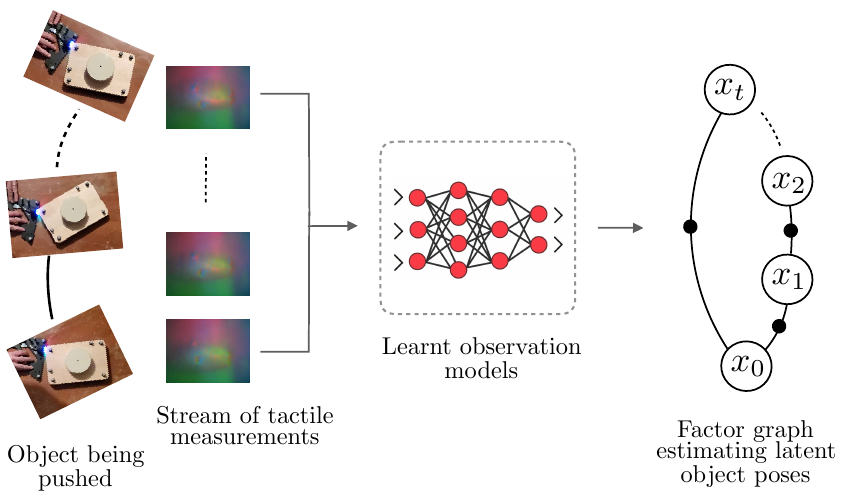}
	\caption{\small Estimating latent object poses from a stream of tactile images modeled as an inference over a factor graph. Since design of observation models for such measurements is challenging, we learn these models to be used as factors within the graph.}
	\label{fig:cover}
	\figGap
\end{figure}

In this paper, we propose learning tactile observation models that are incorporated as factors within a factor graph during inference (Fig. \ref{fig:cover}). The learner takes high-dimensional tactile image measurements and predicts noisy low-dimensional poses that are then integrated by the factor graph optimizer. For the learner, however, accurately predicting object pose directly from a tactile image is typically not possible without additional information such as a tactile map. Our key insight is to instead have the learner predict \textit{relative poses} from image pairs. That is, given a pair of non-sequential tactile images, predict the difference in the pose of the tactile sensor relative to the object in contact. This relative pose information is then used to correct for accumulated drift in the object pose estimate. Observation models trained in this fashion can work for a large class of objects, as long as the local contact surface patches are in distribution to those found in training data.\footnote{While we do not require a tactile map, we do need a category class of the object shape in the observation model and a geometric model prior in the factor graph. These are much easier to obtain as they do not require tactile sensor readings that cover the full object.}

We propose a two-stage approach: we first learn local tactile observation models supervised with ground truth data, and then integrate these along with physics and geometric models as factors within a factor graph optimizer. The tactile observation model is learning-based and integrates with a factor graph. This integration leverages benefits of both learning-based and model-based methods --- wherein, (a) our learnt factor is general and more accurate than engineered functions, and (b) integrating this learnt model within a factor graph lets us reason over a stream of measurements while incorporating structural priors in an efficient, real-time manner. Our main contributions are:
\begin{enumerate}[topsep=0pt]
	\item A novel learnable tactile observation model that integrates as factors within a factor graph.
	\item Tactile factors that work for multiple object shapes.
	\item Real-time object tracking during real-world pushing trials using \textit{only tactile} measurements.
\end{enumerate}

\section{Related Work}
\label{sec:relatedwork}

Localization and state estimation are increasingly solved as smoothing problems given the increased accuracy and efficiency over their filtering counterparts \cite{cadena2016past}. Typically, the smoothing objective is formulated as inference over a factor graph whose variable nodes encode latent states and factor nodes the measurement likelihoods \cite{dellaert2017factor, dellaert2020factor, rosen2021advances}. To incorporate high-dimensional measurements as factors in the graph, one needs an observation model to map between latent states and measurements. These typically are analytic functions e.g. projection geometry for images \cite{mur2017orb, engel2014lsd}, or scan matching for point clouds \cite{dong2019gpu, teixeira2016underwater}. Recent work in visual SLAM has also looked at using such functions on learnt, low-dimensional encodings of the original image measurements \cite{czarnowski2020deepfactors, bloesch2018codeslam}.

While there exist well-studied models within visual SLAM literature, it is hard to obtain such a-priori models for complex sensors such as vision-based tactile sensors \cite{lambeta2020digit, yuan2017gelsight} that capture deformations lacking a straightforward metric interpretation. Recent work is seeking to create increasingly accurate simulator models for such sensors \cite{wang2020tacto, agarwal2020simulation}. Within tactile manipulation literature, one class of approaches using such sensors make use of tactile images directly as feedback to solve various control tasks such as cable manipulation \cite{she2020cable}, in-hand marble manipulation \cite{lambeta2020digit}, box-packing \cite{dong2019tactile}. While efficient for the particular task, it can be difficult to scale across different tasks. An inference module is additionally needed to estimate a common latent state representation like global object poses that can be used for different downstream control and planning tasks.

Prior work on estimating states from touch during manipulation has included filtering methods \cite{izatt2017tracking, saund2017touch, behbahani2016haptic, koval2015mpf, pezzementi2011object, pezzementi2011tactile}, learning-only methods \cite{sundaralingam2019robust, li2014localization}, methods utilizing a prior map or tactile simulators \cite{bauza2019tactile, bauza2020tactile}, and graph-based smoothing methods \cite{yu2018realtime, yu2018realtimesuction, lambert2019joint, suresh2020tactile}. Smoothing approaches that model the problem as inference over a factor graph have the benefits of (a) being more accurate than filtering methods, (b) incorporating structural priors unlike learning-only methods, and (c) recovering global object poses from purely local observation models without needing a global map. Moreover, the graph inference objective can be solved in real-time making use of fast, incremental tools \cite{kaess2008isam, kaess2012isam2, sodhi2020ics} in the literature.

Of these different methods, the work in \cite{yu2018realtime, lambert2019joint, bauza2019tactile, bauza2020tactile} is most closely related. Yu \etal \cite{yu2018realtime}, Lambert \etal \cite{lambert2019joint} solve for a factor graph-based pose estimation objective but use either low-dimensional force readings or engineered functions for high-dimensional readings from sensors like Biotac. Bauza \etal \cite{bauza2019tactile, bauza2020tactile} use different methods for pose estimation but use vision-based tactile sensors most similar to ours. They solve for a global localization objective with either an offline tactile map \cite{bauza2019tactile}, or contact image renderings from a simulator \cite{bauza2020tactile}. We see this global localization objective complementary to the object state tracking objective achieved with our proposed learnt tactile factors.

\begin{figure*}[!t]
	\centering
	\includegraphics[width=0.95\textwidth]{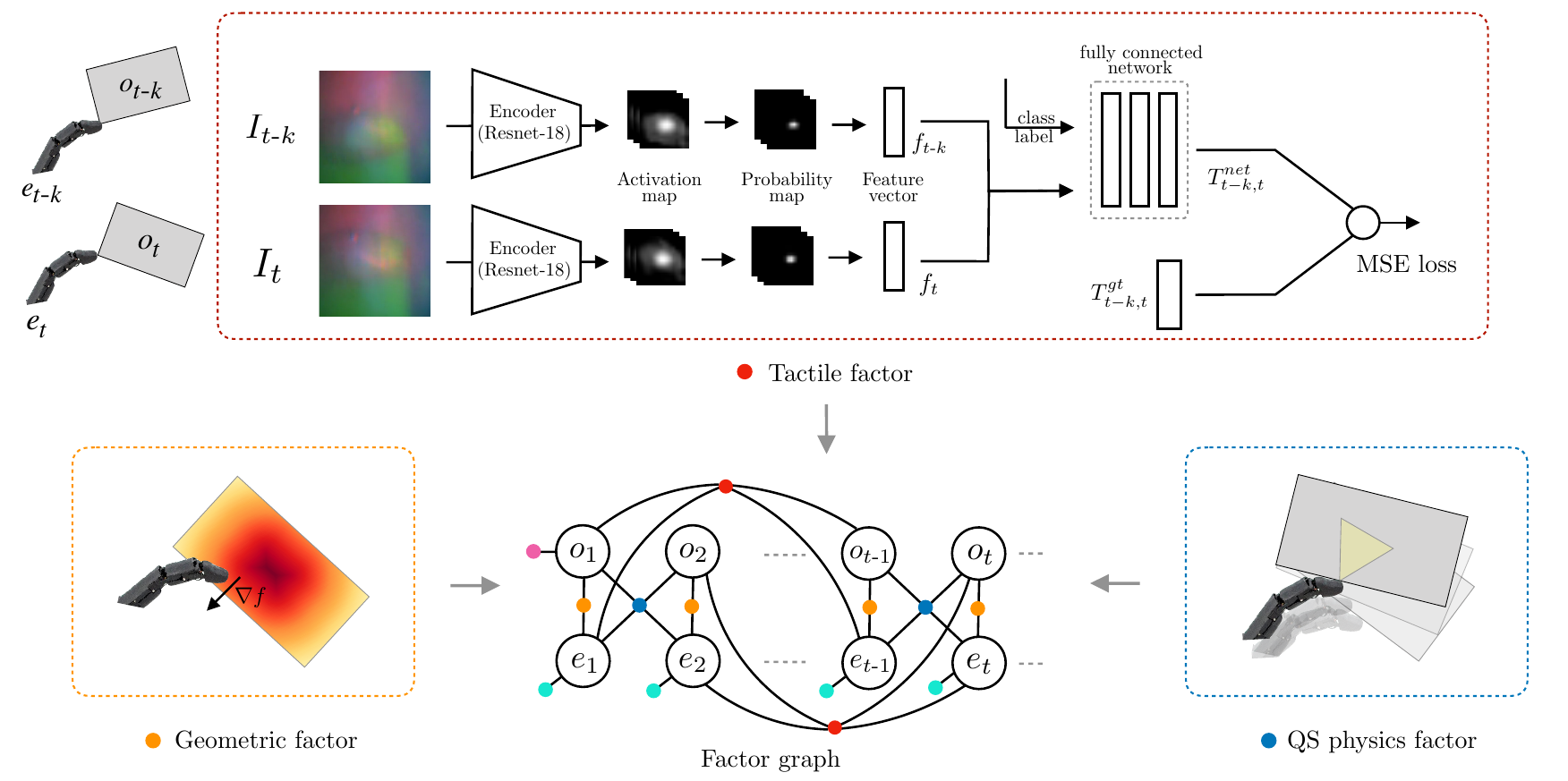}
	\caption{\small Overall approach showing the three factors in the factor graph --- tactile, physics, geometric. Tactile factor (red) predicts relative sensor poses given a pair of tactile images. Physics factors (blue) model quasi-static pushing. Geometric factors (orange) model object and end-effector intersections. We also include global pose priors as unary factors on end-effector (cyan) and the first object pose (pink).}
	\label{fig:approachOverall}
	\figGap
\end{figure*}

\section{Problem Formulation}
\label{sec:probform}
We formulate the estimation problem as inference over a factor graph. A factor graph is a bipartite graph with two types of nodes: variables $x$ and factors $\phi$. Variable nodes are the latent states to be estimated, and factor nodes encode constraints on these variables such as measurement likelihood functions, or physics, geometric models. Maximum a posteriori (MAP) inference over a factor graph involves maximizing the product of all factor graph potentials, i.e.,
\begin{equation}
	\begin{split}
		\label{eq:eq3.1}
		\hat{x}^{}& = \underset{{x}}{\operatorname{argmax}}\ \prod_{i=1}^{m} \phi_i({x}) \\
	\end{split}
\end{equation}

Under Gaussian noise model assumptions, MAP inference is equivalent to solving a nonlinear least-squares problem \cite{dellaert2017factor}. That is, for Gaussian factors $\phi_i({x})$ corrupted by zero-mean, normally distributed noise,
\begin{equation}
	\begin{split}
		\label{eq:eq3.2}
		&\phi_i({x})\propto \exp\left\{-\frac{1}{2}||F_i(x)||_{\Sigma_i}^2\right\} \\
		\Rightarrow \ & \hat{x}=\underset{{x}}{\operatorname{argmin}}\ \frac{1}{2}\sum_{i=1}^{m}||F_i(x)||_{\Sigma_i}^2 \\
	\end{split}
\end{equation}
where, $F_i({x})$ are cost functions defined over states $x$ and include measurement likelihoods or priors derived from physical or geometric assumptions. $||\cdot||_{\Sigma_i}$ is the Mahalanobis distance with covariance $\Sigma_i$.

For our planar pushing setup, states in the graph are the planar object and end-effector poses at every time step $t=1\hdots T$, i.e. $x_t=[o_t\ e_t]^T$, where $o_t, e_t \in SE(2)$. Factors in the graph incorporate tactile observations $F_{tac}(\cdot)$, quasi-static pushing dynamics $F_{qs}(\cdot)$, geometric constraints $F_{geo}(\cdot)$, and priors on end-effector poses $F_{eff}(\cdot)$.

At every time step, new variables and factors are added to the graph. Writing out Eq. \ref{eq:eq3.2} for our setup at time step $T$,
\begin{equation}
	\begin{split}
		\label{eq:eq3.3}
		\hat{x}_{1:T} = & \underset{x_{1:T}}{\operatorname{argmin}}\sum_{t=1}^{T} \left\{\right.||F_{qs}(o_{t-1}, o_{t}, e_{t-1}, e_{t})||^2_{\Sigma_{qs}} + \\
		& ||F_{geo}(o_{t}, e_{t})||^2_{\Sigma_{geo}} + ||F_{tac}(o_{t-k}, o_{t}, e_{t-k}, e_{t})||^2_{\Sigma_{tac}} + \\ & ||F_{eff}(e_{t})||^2_{\Sigma_{eff}} \left.\right\}
	\end{split}
\end{equation}

Eq. \ref{eq:eq3.3} is the optimization objective that we must solve for every time step. Instead of resolving from scratch every time step, we make use of efficient, incremental solvers such as iSAM2 \cite{kaess2012isam2} for real-time inference. Individual cost terms in Eq. \ref{eq:eq3.3} are described in more detail in Section \ref{subsec:approachGraphOptim}.

\section{Approach}
\label{sec:approach}

We present a two-stage approach: we first learn local tactile observation models from ground truth data (Section \ref{subsec:approachObsModel}), and then integrate these models along with physics and geometric models as factors within a factor graph (Section \ref{subsec:approachGraphOptim}). Fig. \ref{fig:approachOverall} illustrates our overall approach showing the three factors, tactile, physics and geometric, being integrated into the factor graph.

\subsection{Tactile observation model}
\label{subsec:approachObsModel}

The goal of learning a tactile observation model is to derive the tactile factor cost term $||F_{tac}(o_{t-k}, o_{t}, e_{t-k}, e_{t})||^2_{\Sigma_{tac}}$ in Eq. \ref{eq:eq3.3} to be used during graph optimization. We do this by predicting a relative transformation and penalizing deviations from this prediction. The relative transformation is that of the sensor (or end-effector) pose relative to the object in contact.

Our learnt tactile observation model consists of a transform prediction network that: Given a pair of non-sequential tactile image inputs $\{I_{t\text{-}k}, I_{t}\}$ at times $\{t\text{-}k, t\}$, predicts the relative transformation $T^{net}_{t\text{-}k,t}$. This is done by featurizing each image. 

\subsubsection*{Feature learning}
For encoding image $I_t$ as feature $f_t$, we use an auto-encoder with a structural keypoint bottleneck proposed in \cite{lambeta2020digit}. It consists of an encoder and decoder using ResNet-18 as the backbone network. The encoder processes image input into a feature map from which $K$ 2D keypoint locations corresponding to maximum feature activations are extracted. At decoding, a Gaussian blob is drawn on an empty feature map for each extracted keypoint. The decoder takes these as inputs and produces a reconstructed image. The auto-encoder is trained in a self-supervised manner with L2 image reconstruction loss along with auxiliary losses optimizing for sparse, non-redundant keypoints.

\subsubsection*{Transform prediction network}
Once we have a trained feature encoder, we use that encoder (with weights fixed) in the transform prediction network to map input image pairs $\{I_{t\text{-}k}, I_{t}\}$ into keypoint feature vectors $\{f_{t\text{-}k}, f_{t}\}$. This is followed by a fully-connected regression network that predicts a relative 2D transformation $T^{net}_{t\text{-}k,t}$ between times $\{t\text{-}k, t\}$ (Fig. \ref{fig:approachOverall}). To make the same transform prediction network work across object classes, we also pass in a one-hot class label vector $c \in \{0,1\}^{N_{c}}$. We expand the feature inputs via an outer product with the class vector, $f_{t-k}\otimes c$, $f_t\otimes c$, and pass this expanded input to the fully-connected layers. The label vector $c$ corresponds to a category class of the object shape, we assume we know this for the object being pushed.

The transform prediction network is trained using a mean-squared loss (MSE) against ground truth data $T^{gt}_{t\text{-}k,t}$. We make the loss symmetric so that the network learns both regular and inverse transform, that is,
\begin{equation}
	\begin{split}
		\label{eq:eq4.a.1}
		MSE = \frac{1}{|\mathcal{S}|}\sum^{}_{(t\text{-}k, t)\in\mathcal{S}} \underbrace{{(T^{gt}_{t\text{-}k, t}-T^{net}_{t\text{-}k, t})}^2}_{\text{transform error}} + \underbrace{(T^{gt}_{t,t\text{-}k}-T^{net}_{t,t\text{-}k})^2}_{\text{inverse transform error}}
	\end{split}
\end{equation}
Here, $T^{gt}_{t\text{-}k, t}$ is the relative transform between end-effector poses at time steps $\{t\text{-}k, t\}$ in object coordinate frame. That is, $T^{gt}_{t\text{-}k, t}=[o_t^{-1}e_t]^{-1}[o_{t\text{-}k}^{-1}e_{t\text{-}k}]$, where $o_t, o_{t\text{-}k}, e_t, e_{t\text{-}k} \in SE(2)$ are ground truth object and end-effector poses obtained using a motion capture system. $\mathcal{S}$ is the set of all non-sequential image pairs $(t\text{-}k, t)$ over a chosen time window $k\in[k_{min}, k_{max}]$.

\subsection{Factor graph optimization}
\label{subsec:approachGraphOptim}

Once we have the learnt local tactile observation model, we integrate it along with physics and geometric models as factors within a factor graph. The factor graph optimizer then solves for the joint objective in Eq. \ref{eq:eq3.3}. Here we look at each of the cost terms in Eq. \ref{eq:eq3.3} in more detail.

\subsubsection*{Tactile factor} 
Measurements from tactile sensor images are incorporated as relative tactile factors denoted by the cost $||F_{tac}(\cdot)||^2_{\Sigma}$ in Eq. \ref{eq:eq3.3}. Our relative tactile factor is a quadratic cost penalizing deviation from a predicted value, that is,
\begin{equation}
	\begin{split}
		\label{eq:eq4.b.1.1}
		||F_{tac}(o_{t-k}, o_{t}, e_{t-k}, e_{t})||^2_{\Sigma_{tac}}:=||{T}^{graph}_{t-k, t}\ominus T^{net}_{t-k, t}||^{2}_{\Sigma_{tac}}
	\end{split}
\end{equation}
where, $T^{net}_{t-k, t}$ is the \textit{predicted} relative transform from the transform prediction network that takes as inputs tactile image measurement at time steps $\{t{\text{-}}k, t\}$. ${T}^{graph}_{t-k, t}$ is the \textit{estimated} relative transform using current variable estimates in the graph. $\ominus$ denotes difference between two manifold elements. $k\in[k_{min}, k_{max}]$ is the time step window over which these relative tactile factors are added. We choose this to be some subset of the training window set $|\mathcal{S}|$ in Eq. \ref{eq:eq4.a.1}.

${T}^{graph}_{t-k, t}$ computed using graph variable estimates $\{o_{t-k}, e_{t-k}, o_{t}, e_{t}\} \in SE(2)$, and $T^{net}_{t-k, t}$ computed as transform prediction network output can be expressed as,
\begin{equation}
	\begin{split}
		\label{eq:eq4.b.1.2}
		{T}^{graph}_{t-k, t} & = [o_{t-k}^{-1}e_{t-k}]^{-1}[o_{t}^{-1}e_{t}] \\
		{T}^{net}_{t-k, t} & \leftarrow \phi^{net}(I_{t-k}, I_{t})
	\end{split}
\end{equation}

\subsubsection*{Quasi-static physics factor}
To model object dynamics as it is pushed, we use a quasi-static physics model. The quasi-static approximation assumes end-effector trajectories executed with negligible acceleration, i.e. the applied pushing force is just large enough to overcome friction without imparting an acceleration \cite{lynch1992manipulation}.

We use the velocity-only quasi-static model from \cite{zhou2017fast} that uses a convex polynomial to approximate the limit surface for force-motion mapping. For sticking contact, the contact point pushing velocity must lie inside the motion cone, resulting in the following relationship between object and contact point velocities:
\begin{equation}
	\begin{split}
		\label{eq:eq4.b.2.1}
		DV = V_p & \\
		\Rightarrow \begin{bmatrix} 1 & 0 & -p_y \\ 0 & 1 & p_x \\ -p_y & p_x & -c^2 \end{bmatrix} \begin{bmatrix}v_x \\ v_y \\ \omega\end{bmatrix} & = \begin{bmatrix}v_{px} \\ v_{py} \\ 0\end{bmatrix}
	\end{split}
\end{equation}
where, $V$ is the object twist, $V_p$ is the contact point velocity, and $c=\tau_{max}/f_{max}$ is a hyper-parameter dependent on pressure distribution of the object \cite{lynch1992manipulation}. We calculate this value assuming pressure to be distributed either uniformly or at the corners/edges of the objects.

The dynamics in Eq. \ref{eq:eq4.b.2.1} is incorporated as the quadratic quasi-static cost term $||F_{qs}(\cdot)||^2_{\Sigma}$ in Eq. \ref{eq:eq3.3}. Expanding this,
\begin{equation}
	\begin{split}
		\label{eq:eq4.b.2.2}
		||F_{qs}(o_{t-1}, o_{t}, e_{t-1}, e_{t})||^2_{\Sigma_{qs}}:=||DV-V_p||^2_{\Sigma_{qs}}
	\end{split}
\end{equation}
Object twist $V$ is computed using object poses $o_{t-1}, o_{t}$ and contact point velocity $V_p$ is computed using end-effector contact point estimate $p_t = \begin{bmatrix}e^x_t & e^y_t\end{bmatrix}$. That is,
\begin{equation}
	\begin{split}
		\label{eq:eq4.b.2.3}
		V = \begin{bmatrix}v_x \\ v_y \\ \omega\end{bmatrix} = \mathcal{R}(o_t)
		\begin{bmatrix}(o^x_t-o^x_{t-1})/dt \\ (o^y_t-o^y_{t-1})/dt \\ (o^{\theta}_t-o^{\theta}_{t-1})/dt\end{bmatrix}
	\end{split}
\end{equation}
\begin{equation}
	\begin{split}
		\label{eq:eq4.b.2.4}
		V_p = \begin{bmatrix}v_{px} \\ v_{py} \\ 0\end{bmatrix} = \mathcal{R}(o_t)
		\begin{bmatrix}(e^x_t-e^x_{t-1})/dt \\ (e^y_t-e^y_{t-1})/dt \\ 0 \end{bmatrix}
	\end{split}
\end{equation}
where, $\mathcal{R}(o_t)$ rotates object and contact point velocities into current object frame.

\begin{figure}[!b]
	\centering
	\includegraphics[width=0.9\columnwidth]{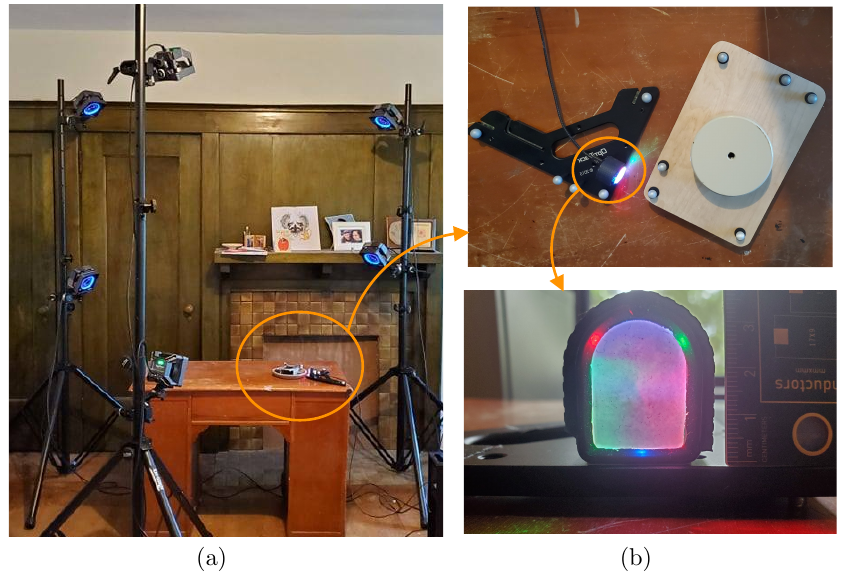}
	\caption{\small Experimental setup \textbf{(a)} shows overall planar pushing setup with motion capture \textbf{(b)} shows closeups of the Digit tactile sensor on an end-effector and the end-effector pushing a rectangular object.}
	\label{fig:experimentalSetup}
\end{figure}

\begin{figure}[!t]
	\centering
	\includegraphics[width=0.9\columnwidth]{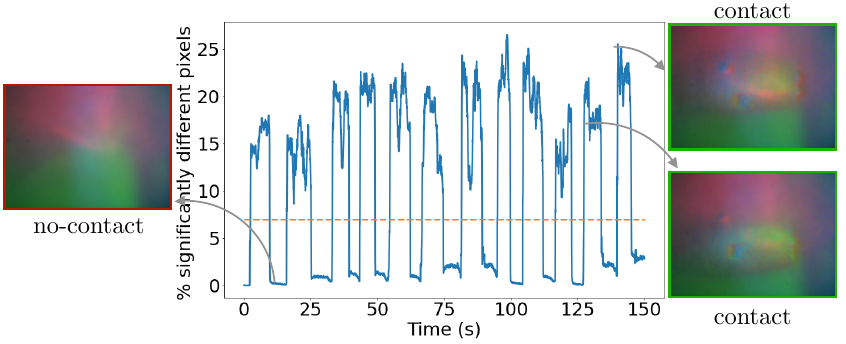}
	\caption{\small Contact within an image is detected if \% of significantly ($\textgreater4\sigma$) different pixels exceed a threshold.}
	\label{fig:contactDetection}
	\figGap
\end{figure}

\begin{figure}[!b]
	\centering
	\includegraphics[width=0.95\columnwidth]{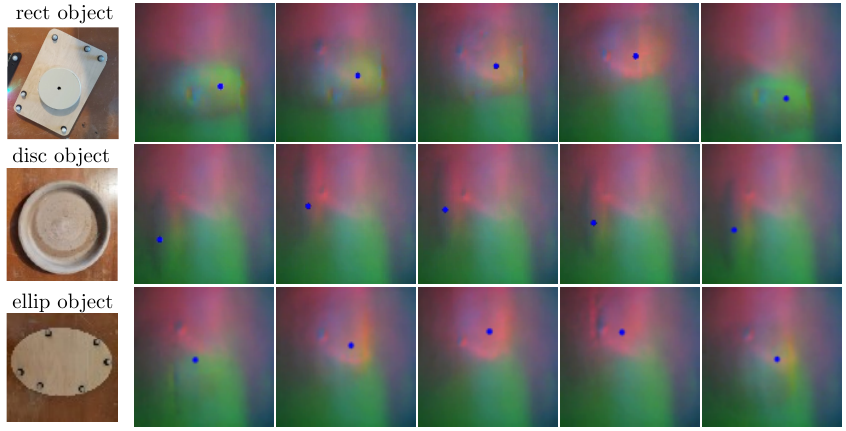}
	\caption{\small Learnt keypoint features in blue tracking the local contact patch in tactile images for \textbf{(top)} rect, \textbf{(middle)} disc and \textbf{(bottom)} ellip objects.}
	\label{fig:keypointFeatures}
\end{figure}

\begin{figure*}[!t]
	\centering
	\includegraphics[width=0.952\textwidth]{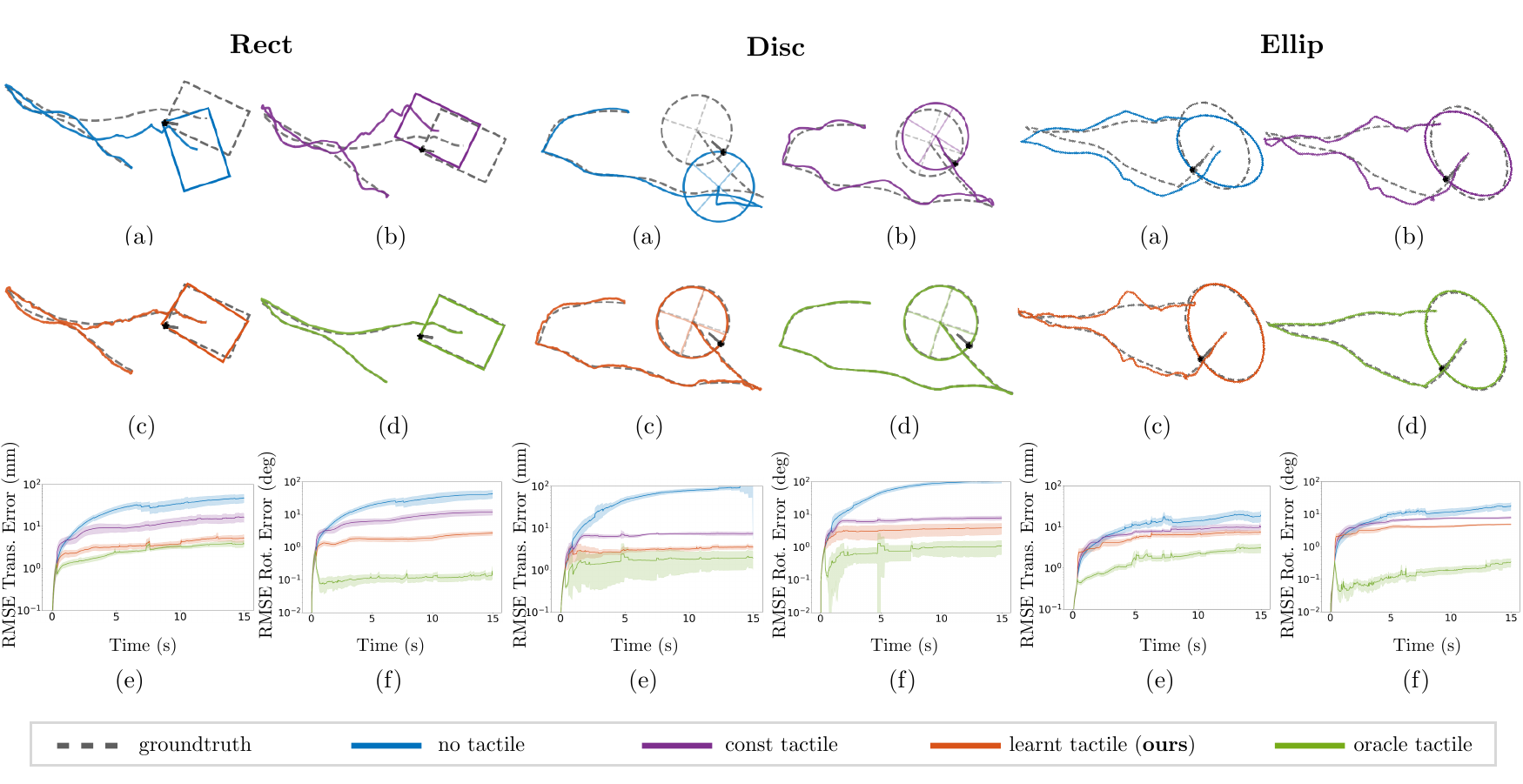}
	\caption{\small Object pose tracking errors over time. \textbf{(a)-(d)} show qualitative tracking results, \textbf{(e), (f)} show quantitative mean-variance plots (semi-log scale) for 50 pushing trials each with rect, disc, ellip objects.}
	\label{fig:estimationResultsTime}
	\figGap
\end{figure*}

\subsubsection*{Geometric factor}
We would like to add a geometric constraint to ensure that the contact point lies on the object surface. We do so as an {intersection cost} between the end-effector and the object. This is similar to the obstacle avoidance cost in \cite{mukadam2018continuous,ratliff2009chomp} but instead of a one-sided cost function, we use a two-sided function to penalize the contact point from lying on either side of the object surface.

To compute this intersection cost, we transform end-effector contact point $p_t = \begin{bmatrix}e^x_t & e^y_t\end{bmatrix}\in\mathbb{R}^2$ into current object frame as $o_{t}^{-1}p_{t}$, and look up its distance value in a precomputed 2D signed distance field map $\xi^{sdf}(\cdot)$ of the object centered around the object. This is incorporated as the quadratic geometric factor cost term $||F_{geo}(\cdot)||^2_{\Sigma}$ in Eq. \ref{eq:eq3.3}. Expanding this,
\begin{equation}
	\begin{split}
		\label{eq:eq4.b.3.1}
		||F_{geo}(o_{t}, e_{t})||^2_{\Sigma_{geo}}:=||\xi^{sdf}(o_{t}^{-1}p_{t})||^2_{\Sigma_{geo}}
	\end{split}
\end{equation}

\subsubsection*{End-effector priors}
Finally we also model uncertainty about end-effector locations as unary pose priors on the end-effector variables. These priors currently come from motion capture readings with added noise, but for a robot end-effector, these would instead come from the robot kinematics.

The pose priors are incorporated as the quadratic end-effector factor cost term $||F_{eff}(\cdot)||^2_{\Sigma}$ in Eq. \ref{eq:eq3.3}. Expanding,
\begin{equation}
	\begin{split}
		\label{eq:eq4.b.3.1}
		||F_{eff}(e_{t})||^2_{\Sigma_{eff}}:=||e_t \ominus \tilde{e}^{mc}_t||^2_{\Sigma_{eff}} \\ 
	\end{split}
\end{equation}
where, $\tilde{e}^{mc}_t={e}^{mc}_t\oplus\ \mathcal{N}(0,\Sigma_{eff})$ are poses from the motion capture system with added Gaussian noise.

\section{Results and Evaluation}
We evaluate our approach qualitatively and quantitatively on a number of real-world planar pushing trials where the pose of an object is unknown and must be estimated. We compare against a set of baselines on metrics like learning errors, estimation accuracy and runtime performance. Learnt tactile factors are trained using PyTorch \cite{paszke2019pytorch}. These, along with engineered physics and geometric factors, are incorporated within GTSAM C++ library \cite{dellaert2012factor}. We use the iSAM2 \cite{kaess2012isam2} solver for efficient, incremental optimization.

\subsection{Experimental setup}

Fig. \ref{fig:experimentalSetup}(a) shows the overall experimental setup for the pushing trials. We use an OptiTrack motion capture system to record ground truth object and end-effector poses for training and evaluating tactile observation models. Fig. \ref{fig:experimentalSetup}(b) shows a closeup of the object, end-effector and the Digit tactile sensor mounted on an end-effector. The Digit sensor provides high-dimensional RGB images of the local deformation at the contact point \cite{lambeta2020digit}.

\subsection{Tactile factor learning}
We now look at performance of the first stage of our approach, i.e. learning tactile observation models.

\subsubsection*{Contact detection}
The first step is to detect contact in tactile images. We do so by first subtracting a mean no-contact image from the current image. If \% of significantly ($\textgreater4\sigma$) different pixels exceeds a threshold, then contact is declared. We found this simple method works reliably on different pushing trials.

\subsubsection*{Keypoint features}
Fig. \ref{fig:keypointFeatures} shows results for $K\text{=}1$ keypoint features learnt using the auto-encoder network described in Sec \ref{subsec:approachObsModel}. The tactile image shows an elliptical contact patch where the curvature varies with local surface geometry of the object. The learnt keypoint features are able to track the patch center over time.

\subsubsection*{Tactile model performance}
Table \ref{tab:tactileModelPerf} compares mean-squared losses (Eq. \ref{eq:eq4.a.1}) on the validation dataset for different choices of the transform prediction network described in Sec. \ref{subsec:approachObsModel}. We use a total dataset of $\sim 3000$ tactile images per object shape with a $50\text{-}50$ train-test split. For training the transform prediction network, we create a dataset of pairwise images over a chosen time window $(t\text{-}k, t)$ with $k\in[10,40]$. Table \ref{tab:tactileModelPerf}, \textit{const} is a zeroth order model that predicts mean relative transform of the training dataset. This is equivalent to using only contact detection information irrespective of contact patch locations in the images. \textit{eng-feat} and \textit{learnt-feat} represent engineered vs learnt keypoint features. For engineered features, we find a least-squares ellipse fit to detected contours in the image, as a generalization to points/line features used in related work \cite{hogan2020tactile}. \textit{linear} and \textit{nonlinear} refer to using linear or nonlinear ReLU activations in the fully-connected layers. We see that models using \textit{learnt-feat} have lower losses over \textit{eng-feat}. We also see \textit{linear} models have lower losses than \textit{const}. The \textit{nonlinear} models don't show a significant improvement over the \textit{linear} models. This is because there is a strong approximately linear relationship between the keypoint feature motions and relative rotation of the sensor about the contact point.


\begin{figure}[!t]
	\centering
	\includegraphics[width=0.9\columnwidth]{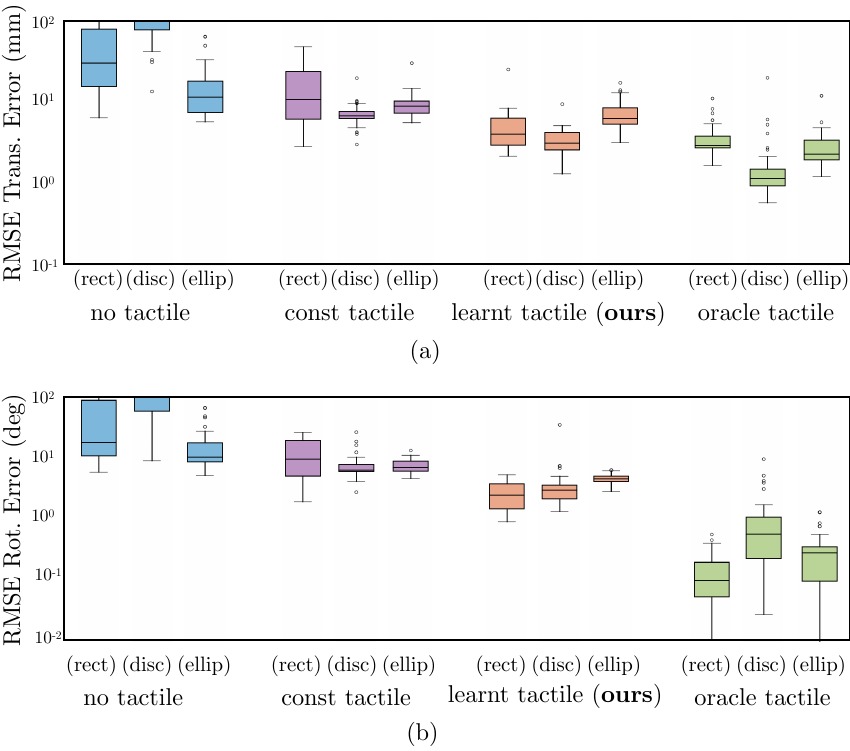}
	\caption{Final estimation error box plots showing \textbf{(a)} translation and \textbf{(b)} rotation error values (semi-log scale).}
	\label{fig:estimationErrorsFinal}
	\figGap
\end{figure}

\begin{figure}[!t]
	\centering
	\includegraphics[width=0.9\columnwidth]{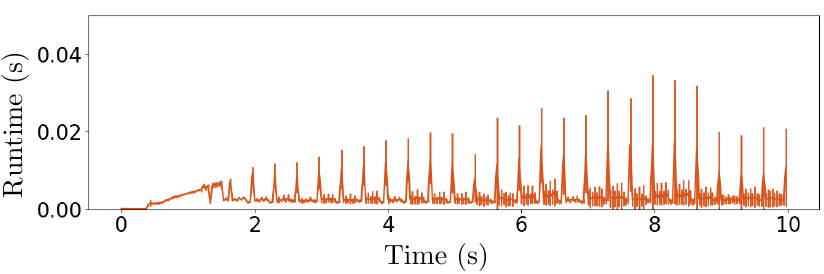}
	\caption{Graph optimizer runtime}
	\label{fig:runtimePerformance}
	\figGap
\end{figure}

\begin{table}[!t]
	\centering
	\caption{\small Tactile Model Performance (Validation loss)}
	\label{tab:tactileModelPerf}
	\setlength{\tabcolsep}{4pt}
	\resizebox{\columnwidth}{!}{
	\begin{tabulary}{\textwidth}{LCCCC} 
		\multirow{3}{*}{\textbf{Model complexity}} & & & \textbf{Datasets} & \\ \cmidrule{2-5}
		& \textbf{Disc} & \textbf{Rect} & \textbf{Ellip} & \textbf{Combined} \\ \midrule
		{const} &  5.5e-3 & 15e-3 & 6.7e-3 & 9.6e-3 \\
		{linear with eng-feat} & 5.1e-3 & 6.7e-3 & 4.2e-3 & 5.9e-3 \\
		{linear with learnt-feat} & \textbf{1.0e-3} & \textbf{1.1e-3} & \textbf{1.7e-3} & \textbf{1.5e-3} \\
		{nonlinear with eng-feat} & 3.0e-3 & 5.2e-3 & 3.3e-3 & 4.2e-3 \\
		{nonlinear with learnt-feat} & \textbf{1.0e-3} & 1.7e-3 & 3.0e-3 & 2.4e-3 \\ \midrule
	\end{tabulary}
	}
	\figGap
\end{table}

\subsection{Factor graph optimization}

We now look at the final task performance of estimating object poses using learnt tactile factors along with physics/geometric factors. For all runs, we use the same linear with learnt-feat tactile model network from Table \ref{tab:tactileModelPerf} trained on the combined datasets. The tactile model network is conditioned on category label $c$ of the object shape being used. $c$ corresponds to different clusters for local curvatures --- disc, ellip, rect-corners and rect-edges. We keep the covariance parameters same across all runs in the graph, i.e. $\Sigma_{tac}\text{=}\diag(1,1,1e\text{-}5)$, $\Sigma_{qs}\text{=}\diag(1e\text{-}3,1e\text{-}3,1e\text{-}3)$, $\Sigma_{geo}\text{=}\diag(1e\text{-}2)$, $\Sigma_{eff}\text{=}\diag(1e\text{-}5,1e\text{-}5,1e\text{-}5)$. First pose of the object is assumed to be known, and added as a unary prior to graph.

\subsubsection*{Qualitative tracking performance}
Fig. \ref{fig:estimationResultsTime}(a) shows that physics and geometric factors alone cause object poses to drift over time. Fig. \ref{fig:estimationResultsTime}(b)-(d) shows tracking with different tactile models: \textit{const}, \textit{learnt}, and \textit{oracle}. \textit{const} predicts a constant relative transform value for each tactile factor. This approach is equivalent to using only a contact detector. While it improves over \textit{no tactile}, it is unable to correct object rotations relative to contact point leading to drifting object pose estimates. This effect is most pronounced in the rect trajectories, where pushing along a corner cause large object rotations about the contact point. \textit{learnt} is our proposed method using the transform prediction network, while \textit{oracle} provides ground truth relative transforms. We see that the \textit{learnt} tactile model recovers object poses close to their true trajectory, and matches oracle performance closely.

\subsubsection*{Quantitative tracking performance}
Fig. \ref{fig:estimationResultsTime}(e),(f) show mean-variance plots of RMSE rotational and translation object pose errors over time. Errors are computed over $50$ pushing sequences each for rect, disc, ellip objects. The sequences have varying pushing trajectories making and breaking contact typically $2\text{-}3$ times. Fig. \ref{fig:estimationErrorsFinal} additionally shows summary statistics of the RMSE translation and rotational errors at the final time step $T=15s$. The learnt model performance is closest to oracle tactile performance recovering true object poses up to ${\sim}10$mm translational errors and ${\sim}5\degree$ rotational errors.

\subsubsection*{Runtime performance}
Finally, Fig. \ref{fig:runtimePerformance} shows runtime per iteration of the graph optimizer. Runtime stays relatively constant with new measurements and priors added every step. 

\section{Discussion}

We presented a factor graph based inference approach for estimating object poses from touch using vision-based tactile sensors. We proposed learning tactile observation models that directly integrate as factors within the graph. We demonstrated that our method is able to reliably track object poses for over 150 real-world planar pushing trials using tactile measurements alone. 

As future work, in the tactile observation model, we would like to learn to model a distribution over relative poses instead of only the mean-squared error values. This should improve performance in cases where the relative pose uncertainty is asymmetric or varies significantly between contact episodes. We would also like to learn richer feature descriptors that describe contact patch geometry in addition to the patch centers. This should allow us to not have to condition on the object category label, and instead capture this information with the feature descriptor itself. Finally, to make these tactile factors work on more complex manipulation tasks, different physics priors will need to be incorporated.

\section*{\footnotesize{Acknowledgements}}
\footnotesize{We would like to thank the DIGIT team, in particular R. Calandra, P.W. Chou, and M. Lambeta for support with the sensor, software and helpful discussions. We would also like to thank Z. Dong, D. Gandhi, and anonymous reviewers for helpful feedback and suggestions.}

\balance




\footnotesize
\bibliographystyle{ieeetr}
\clearpage
\bibliography{references}

\begin{figure*}[!b]
	\centering
	\includegraphics[width=\textwidth]{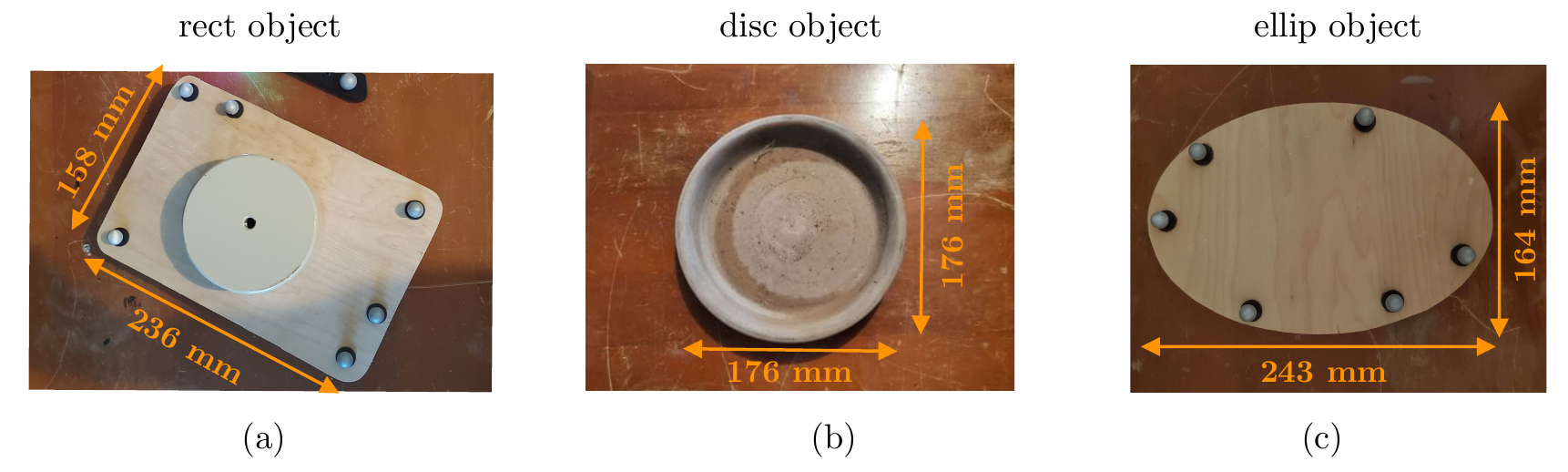}
	\caption{Different objects with dimensions used in planar pushing trials.}
	\label{fig:planarPushingObjects}
\end{figure*}

\newpage

\normalsize
\section*{Appendix}

Here we look at some additional results for object pose tracking during planar pushing tasks. Fig. \ref{fig:planarPushingObjects} shows the three different objects being pushed by the Digit tactile sensor mounted on an end-effector. Objects are pushed over $50$ varying sequences for each object. We perform following variations across the pushing sequences:
\begin{itemize}
\item Trajectories are varied as combinations of straight-line and curved pushes. 
\item Each sequence makes and breaks contact typically 2-3 times.
\item Object is pushed at different locations, e.g. pushing at both edges and corners for the rect object.
\end{itemize}

\begin{figure*}[!t]
	\centering
	\begin{subfigure}[b]{\textwidth}
	\includegraphics[width=\textwidth]{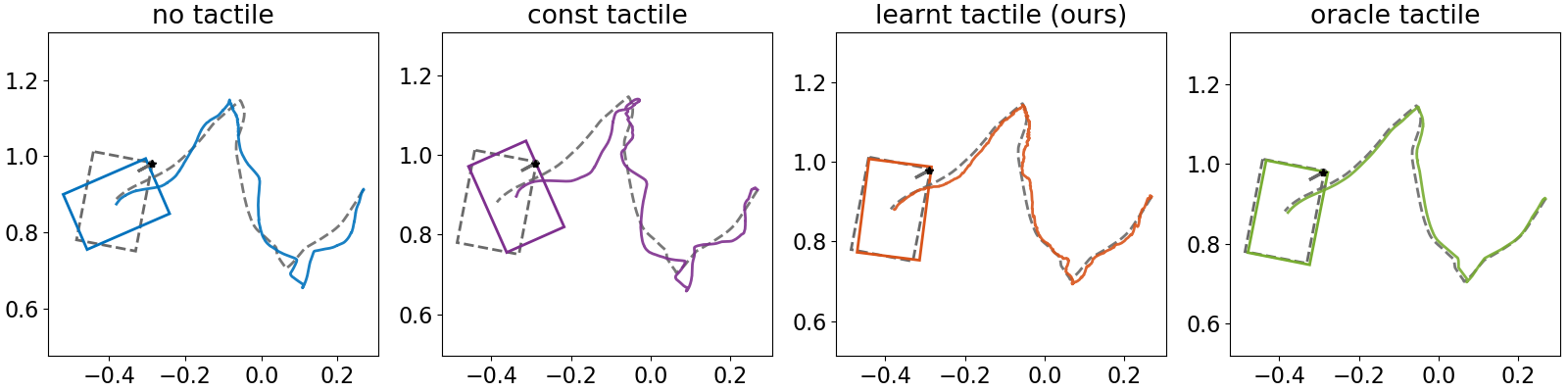}
    \end{subfigure}
	\begin{subfigure}[b]{\textwidth}
	\includegraphics[width=\textwidth]{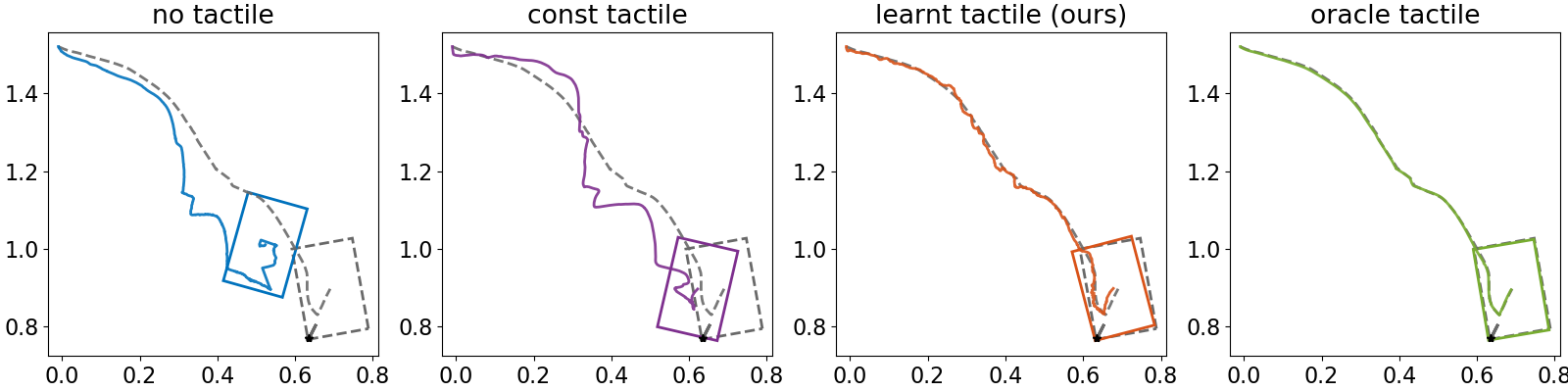}
    \end{subfigure}
	\begin{subfigure}[b]{\textwidth}
	\includegraphics[width=\textwidth]{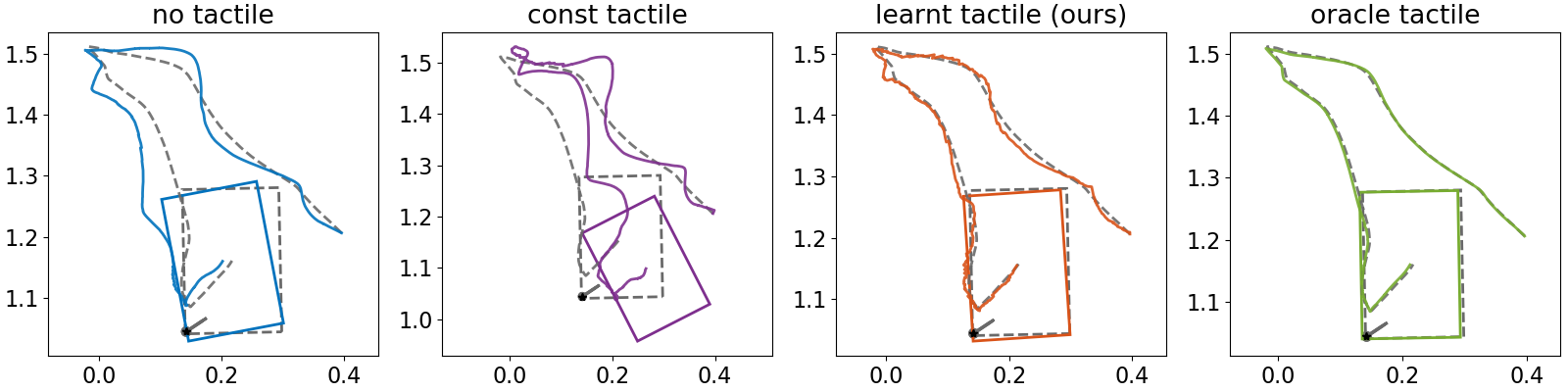}
    \end{subfigure}
	\begin{subfigure}[b]{\textwidth}
	\includegraphics[width=\textwidth]{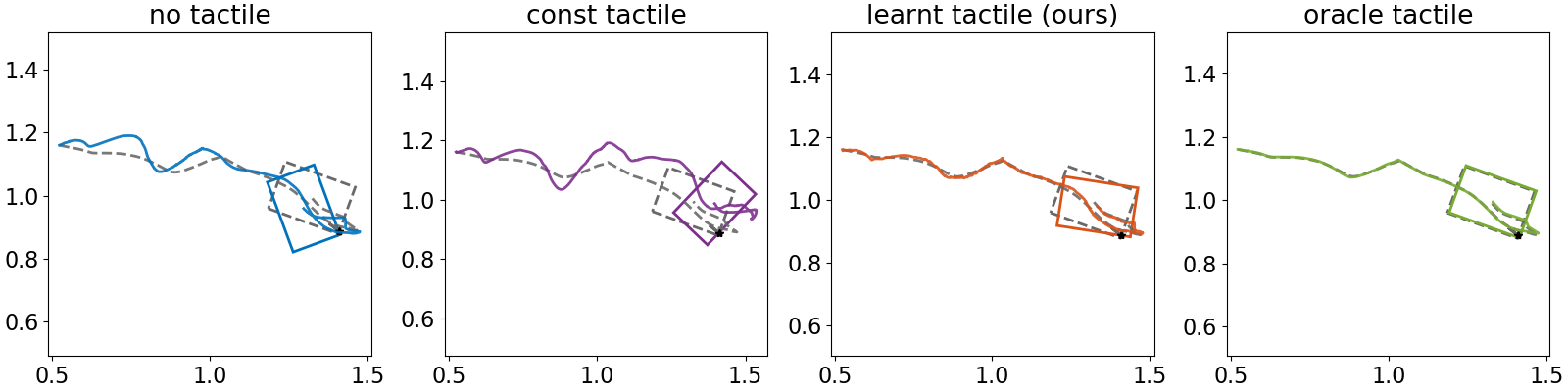}
    \end{subfigure}
	\begin{subfigure}[b]{\textwidth}
	\includegraphics[width=\textwidth]{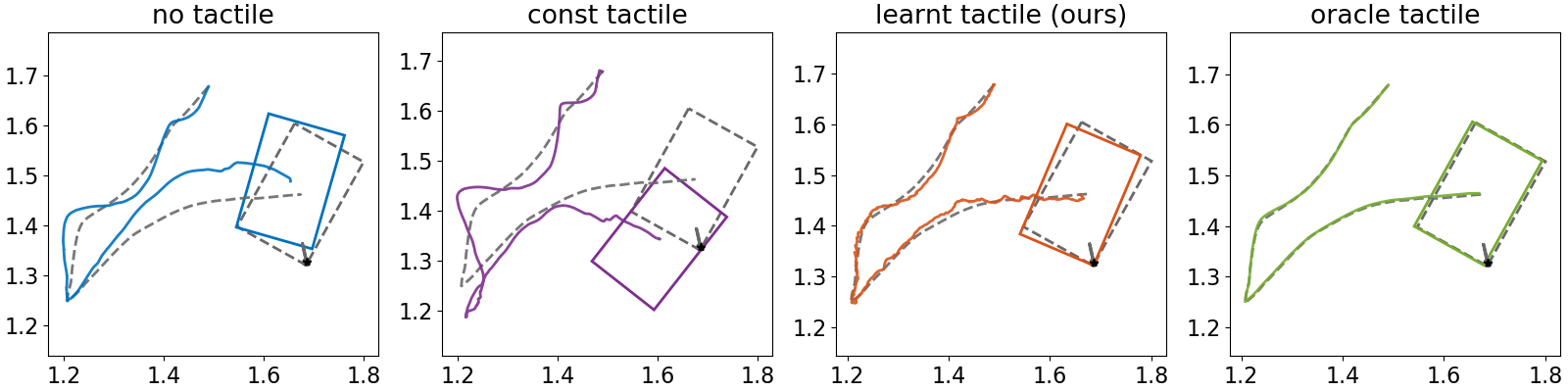}
    \end{subfigure}
	\caption{Rect object pose tracking over time (15s).}
	\label{fig:rectObjectTracking1}
\end{figure*}

\begin{figure*}[!t]
	\centering
	\begin{subfigure}[b]{\textwidth}
	\includegraphics[width=\textwidth]{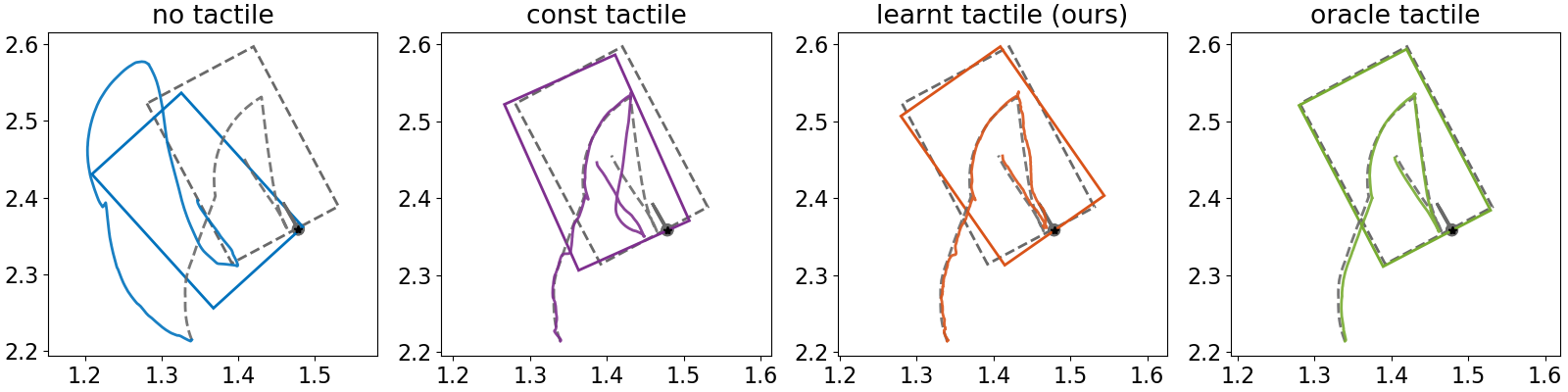}
    \end{subfigure}
	\begin{subfigure}[b]{\textwidth}
	\includegraphics[width=\textwidth]{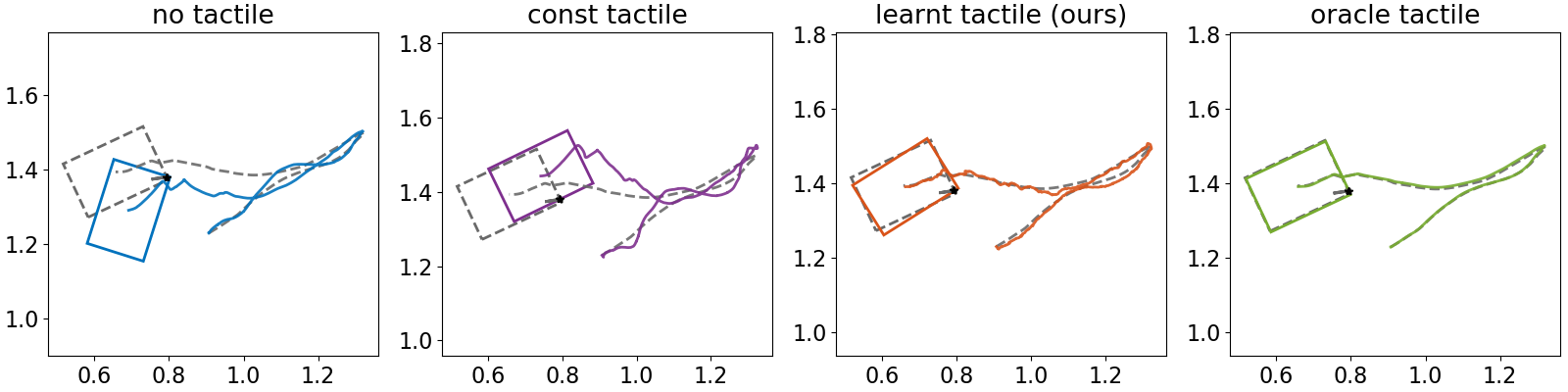}
    \end{subfigure}
	\begin{subfigure}[b]{\textwidth}
	\includegraphics[width=\textwidth]{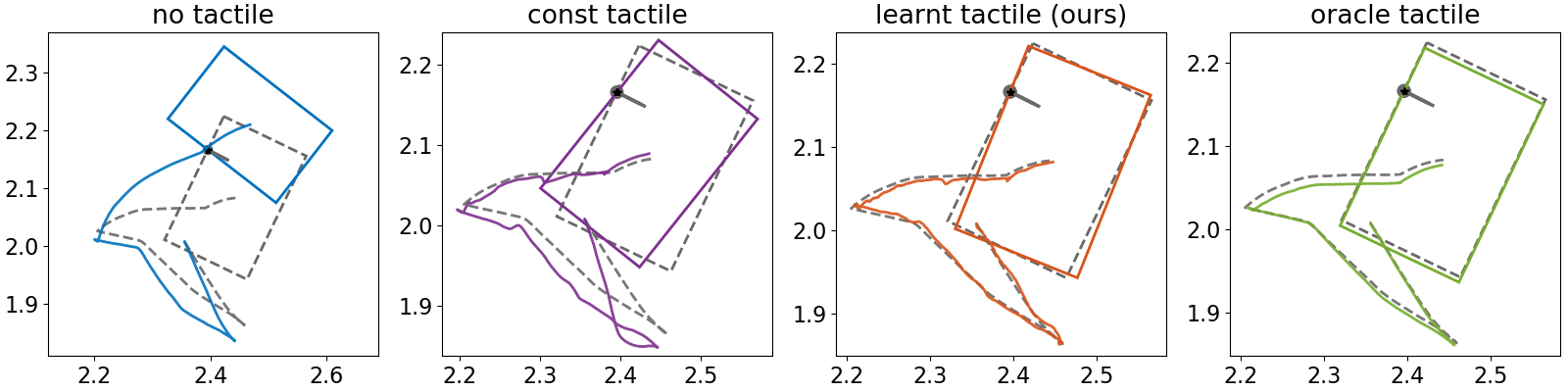}
    \end{subfigure}
	\begin{subfigure}[b]{\textwidth}
	\includegraphics[width=\textwidth]{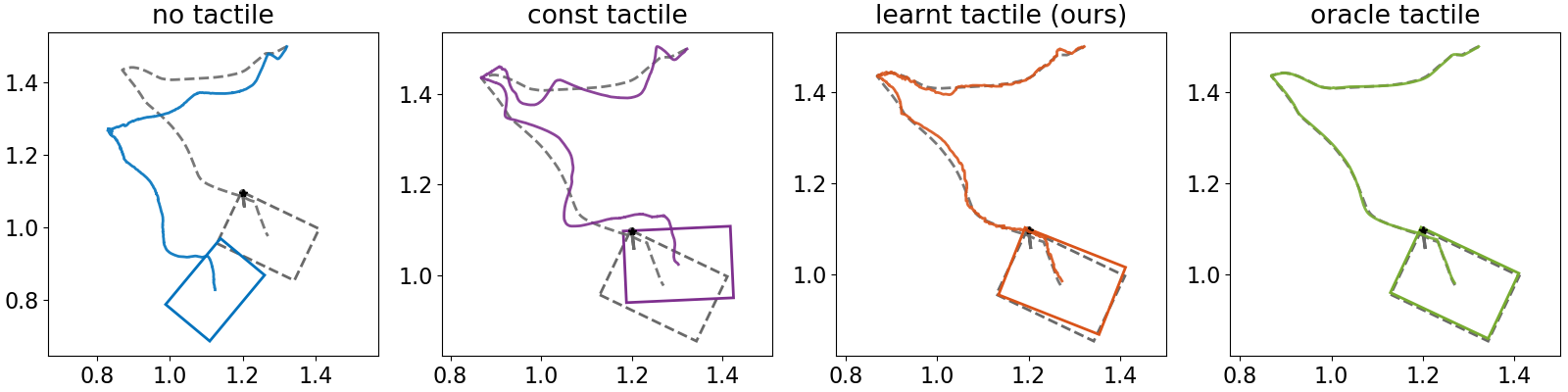}
    \end{subfigure}
	\begin{subfigure}[b]{\textwidth}
	\includegraphics[width=\textwidth]{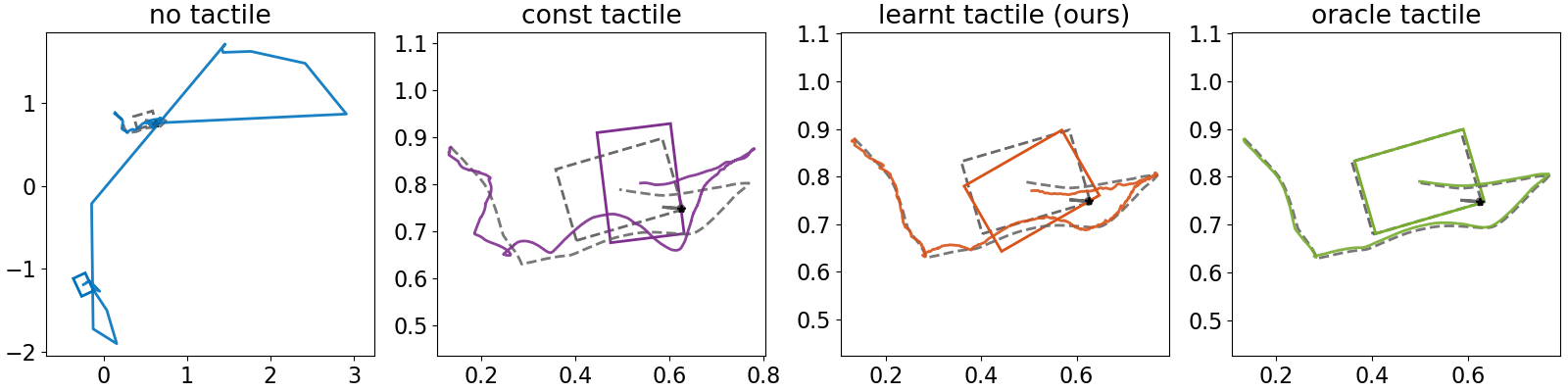}
    \end{subfigure}
	\caption{Rect object pose tracking over time (15s).}
	\label{fig:rectObjectTracking2}
\end{figure*}

\begin{figure*}[!t]
	\centering
	\begin{subfigure}[b]{\textwidth}
	\includegraphics[width=\textwidth]{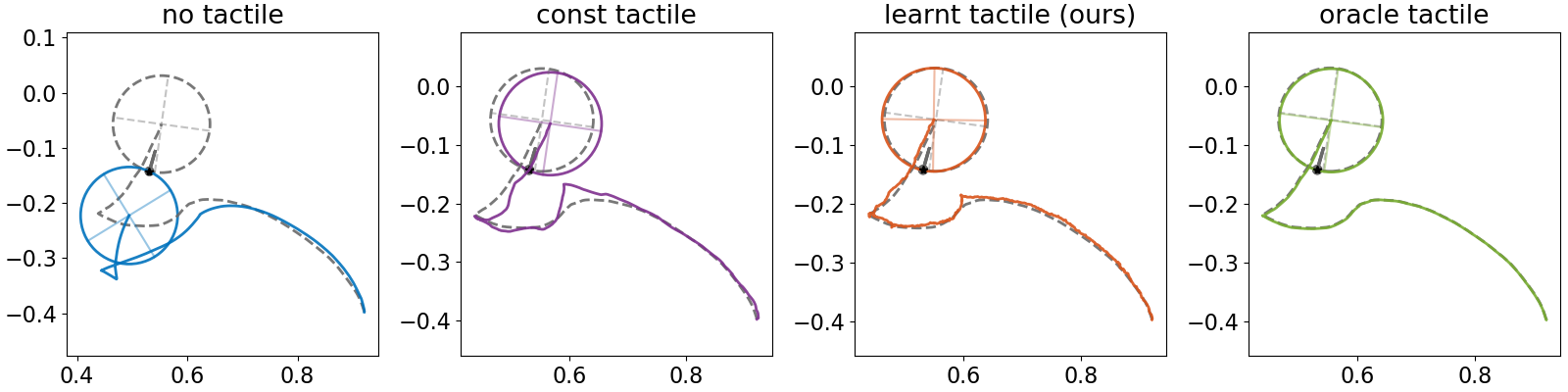}
    \end{subfigure}
	\begin{subfigure}[b]{\textwidth}
	\includegraphics[width=\textwidth]{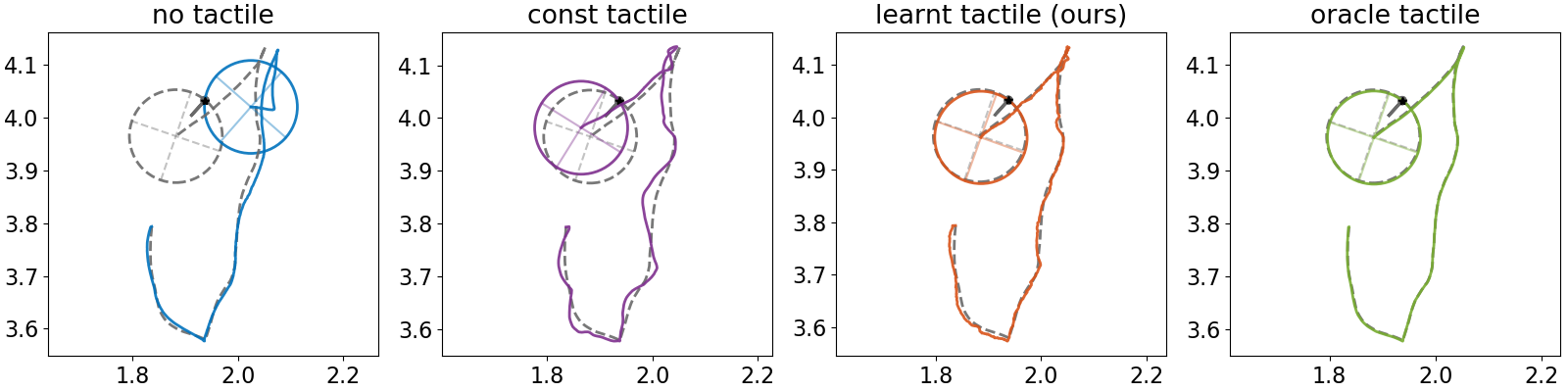}
    \end{subfigure}
	\begin{subfigure}[b]{\textwidth}
	\includegraphics[width=\textwidth]{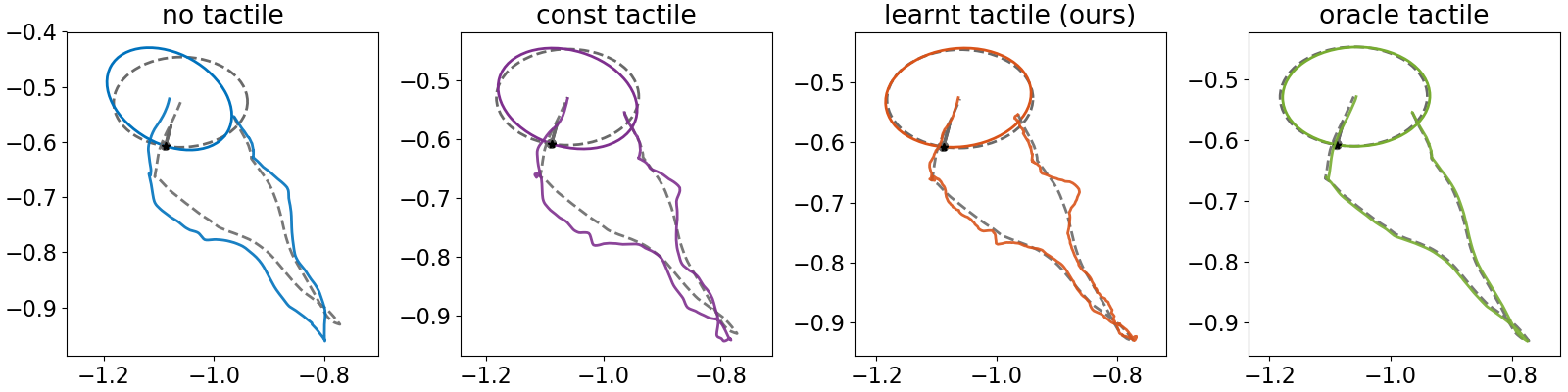}
    \end{subfigure}
	\begin{subfigure}[b]{\textwidth}
	\includegraphics[width=\textwidth]{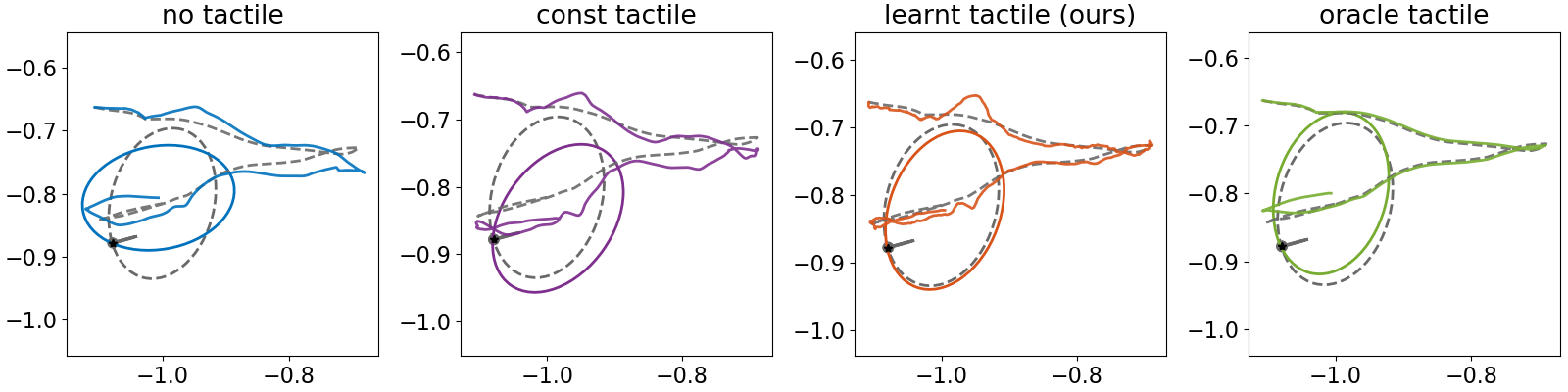}
    \end{subfigure}
	\begin{subfigure}[b]{\textwidth}
	\includegraphics[width=\textwidth]{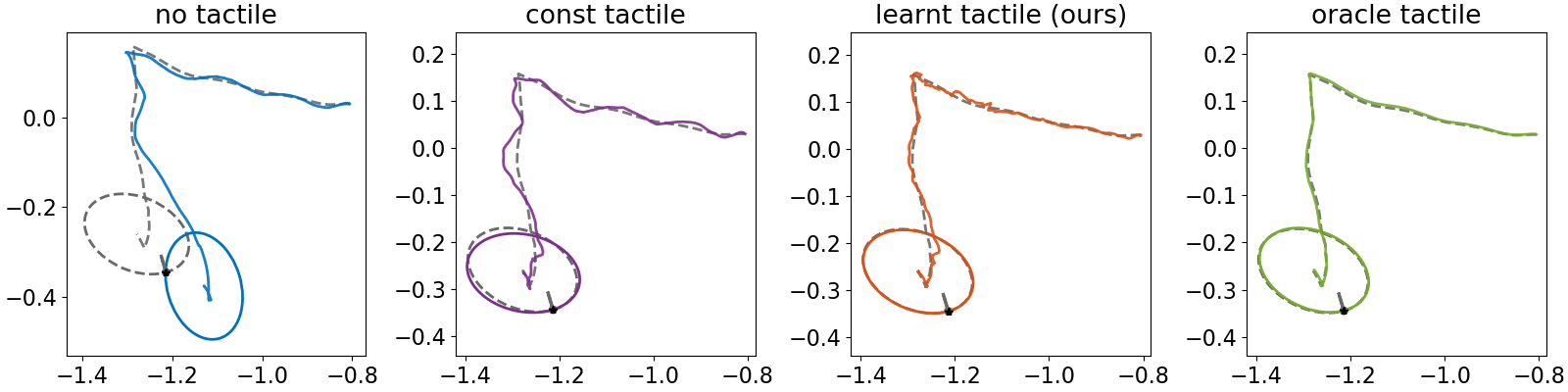}
    \end{subfigure}
	\caption{Disc, Ellip object pose tracking over time (15s).}
	\label{fig:discEllipObjectTracking}
\end{figure*}

Figs. \ref{fig:rectObjectTracking1}, \ref{fig:rectObjectTracking2} show qualitative object tracking results for the rect object, and Fig. \ref{fig:discEllipObjectTracking} for the other objects: disc and ellip over these varying sequences. In the \textit{no tactile} case, use of physics and geometric factors alone cause object poses to drift over time. The last three columns show tracking with different tactile models: \textit{const}, \textit{learnt}, \textit{oracle}. \textit{const} predicts a constant relative transform for each tactile factor. This approach is equivalent to using only a contact detector, i.e. using measurements that a standard non image-based contact sensor would give. While it improves over \textit{no tactile}, it is unable to correct object rotations relative to contact point leading to drifting object pose estimates. This effect is most pronounced in the rect trajectories, where pushing along a corner cause large object rotations about the contact point. \textit{learnt} is our proposed method using the transform prediction network, while \textit{oracle} provides ground truth relative transforms. We see that the \textit{learnt} tactile model recovers object poses close to their true trajectory, and matches oracle performance closely.
 
\end{document}